\documentclass[sigconf]{acmart}
\AtBeginDocument{%
  }

\usepackage{soul}
\usepackage{url}
\usepackage{graphicx}
\usepackage{amsmath}
\usepackage{amsthm}
\usepackage{booktabs}
\usepackage{algorithm}
\usepackage{algorithmic}
\usepackage[switch]{lineno}
\newcommand{\indep}{\perp \!\!\! \perp}
\usepackage{bbm}
\usepackage[normalem]{ulem}
\useunder{\uline}{\ul}{}
\usepackage[table]{xcolor}
\usepackage{booktabs}
\usepackage{enumitem}

\usepackage[font=small]{caption}

\copyrightyear{2026}
\acmYear{2026}
\setcopyright{cc}
\setcctype{by}
\acmConference[CIKM '26]{Proceedings of the 35th ACM International Conference on Information and Knowledge Management}{November 07--11, 2026}{Rome, Italy}
\acmBooktitle{Proceedings of the 35th ACM International Conference on Information and Knowledge Management (CIKM '26), November 07--11, 2026, Rome, Italy}
\acmDOI{10.1145/3799682.3840105}
\acmISBN{979-8-4007-2539-5/2026/11}

\begin{document}

\title{CATeye: Coupled Attribute–Topology Invariance Learning for Voucher Abuse Detection}

\author{Tian Tian}
\email{tian006@e.ntu.edu.sg}
\affiliation{%
  \institution{Nanyang Technological University}
  \country{Singapore}
}

\author{Shuaicheng Niu}
\email{shuaicheng.niu@ntu.edu.sg}
\affiliation{%
  \institution{Nanyang Technological University}
  \country{Singapore}
}

\author{Hao Kuang}
\email{h.kuang@alibaba-inc.com}
\affiliation{
  \institution{Lazada Inc.}
  \country{Singapore}}

\author{Yuanhang Hu}
\email{huyuanhang.hyh@alibaba-inc.com}
\affiliation{
  \institution{Lazada Inc.}
  \city{Beijing}
  \country{China}}

\author{Dong Li}
\email{shiping@taobao.com}
\affiliation{
  \institution{Alibaba Group}
  \city{Hangzhou}
  \state{Zhejiang}
  \country{China}}
  
\author{Zhiqi Shen}
\email{zqshen@ntu.edu.sg}
\affiliation{
  \institution{Nanyang Technological University}
  \country{Singapore}
}
\renewcommand{\shortauthors}{Tian Tian et al.}

\begin{abstract}
Voucher abuse poses a major challenge in e-commerce, where malicious users exploit promotional vouchers for profit. Unfortunately, fraud patterns evolve rapidly over time and across regions, causing distribution shifts that degrade existing detection models unless retrained frequently. To tackle this, we propose the Coupled Attribute–Topology Invariance Learning framework (CATeye). The key challenge arises from coupled attribute–topology shift, where edges built from attribute proximity cause environment-driven attribute shift to induce shifted topology, thereby amplifying variant signals through GNN message passing. CATeye \textit{sees through} such coupled shifts with two learnable selectors. First, an Attribute Invariance Selector (AIS) learns node-adaptive masks to filter out non-invariant attributes. Then, conditioned on retained invariant attributes, an Edge Invariance Selector (EIS) samples an invariant subgraph and isolates non-invariant edges. Using the resulting invariant and non-invariant components, CATeye constructs multiple views and applies view-specific objectives to emphasize domain-invariant representations while suppressing domain-specific variations. Experiments on both a proprietary dataset from Lazada, a major Southeast Asian e-commerce platform, and a public benchmark show that CATeye consistently outperforms nine strong domain generalization and graph anomaly detection baselines, achieving up to an 8.61\% improvement in average F1 score over the strongest baseline. Source code is publicly available at \url{https://github.com/Tian0426/CATeye}.
\end{abstract}


\maketitle

\section{Introduction}

E-commerce platforms widely distribute promotional vouchers to acquire users~\cite{he2017understanding,cao2021big}, but this creates significant financial risk. Malicious actors register fake accounts to redeem vouchers, purchase goods for resale, or collude with sellers. Such behaviors cause substantial financial losses and undermine fairness for legitimate users, making accurate and robust voucher abuse detection essential. 

Voucher abuse often occurs in \textit{coordinated groups}, where multiple fraudulent accounts share common entities (e.g., devices, emails). Conventional heuristic rules based on shared entities~\cite{del2017enhancing} quickly become ineffective as abusers adapted their strategies. To overcome the rigidity of heuristics, subsequent machine learning–based methods operate on tabular data extracted from orders to capture statistical and behavioral patterns~\cite{alonge2021enhancing}. However, they miss the essential inter-order relations. To address this, the graph-based VPGNN~\cite{wen2023voucher} models inter-order relations with graph neural networks (GNNs). Despite its promising performance, it relies on frequent fine-tuning with labeled data to handle distribution shifts in new time periods or markets, as discussed below.

In practice, voucher abuse is highly dynamic. First, it \textbf{evolves rapidly over time} as abusers continuously adapt to deployed detection models. Emerging methods to bypass detection based on common devices or addresses include using randomized virtual devices and obfuscating delivery addresses. Second, abuse behavior \textbf{varies substantially across regions}. A device type or payment method strongly correlated with abuse in one country may be common among legitimate users in another country. As a result, models trained on a specific time period or country often degrade when applied to new periods or regions unless retrained frequently. In reality, retraining on every new market or campaign cycle is operationally prohibitive given the scale and pace of promotions. These challenges highlight the need for domain generalization (DG), which aims to train models that generalize to unseen target environments without fine-tuning. 

\begin{figure}[t]
  \centering
  \includegraphics[width=0.7\linewidth]{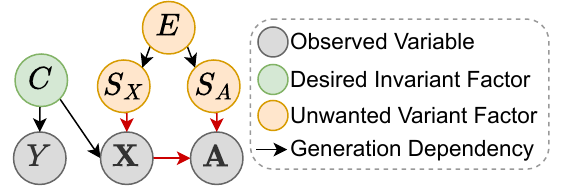}
  \caption{Illustrative model of graph generation under coupled shifts. Environment $E$ introduces variation into attributes $X$ and structure $A$ via $S_X$ and $S_A$, while the invariant factor $C$ drives labels $Y$. The observed $X$ and $A$ are generated through $\{C,S_X\}\!\to\!X$ and $\{X,S_A\}\!\to\!A$. Arrows indicate generation dependencies. Mechanisms causing coupled shifts are highlighted in red. }
  \label{fig: scm}
\end{figure}

Voucher abuse graphs exhibit \textit{coupled attribute-topology shifts} that make generalization especially challenging. In voucher graphs, shifts arise in both attributes and structure. For example, certain device types or payment methods correlate with abuse only in specific regions, and coincidental connections emerge from overlapping purchasing behaviors during large-scale promotional campaigns, resulting in variant attributes and connections that harm generalization. Moreover, variant attributes can further induce environment-dependent structural shifts, since edges are constructed from attribute proximity. Under such coupled shifts, GNN message passing propagates variant signals through shift-induced edges, amplifying domain-specific patterns that do not generalize. 

This unique coupled shift challenges existing domain generalization methods. To address distribution shift, recent advances in graph DG aim to learn representations that remain invariant across source domains~\cite{zhang2024survey}. EERM~\cite{wu2022handling} and CaNet~\cite{wu2024graph} enforce invariance in the embedding space via risk extrapolation on virtual environments or environment estimation. Data-centric approaches such as TRACI~\cite{zhao2025traci} enhance cross-domain alignment through topology-aware adversarial perturbations and prototypical mixup. 
However, these approaches assume that node attributes and graph structure vary independently, or that invariance can be fully enforced in the embedding space. Both assumptions \textit{fail} under coupled shifts, where non-invariant attributes induce shifted edges that GNN message passing then amplifies. This makes existing methods \textit{insufficient} for robust generalization in voucher abuse detection. 
Motivated by this, we explicitly isolate non-invariant attributes and shift-induced edges to learn representations that remain reliable under coupled shifts across time periods and regions. 

To this end, we propose \textbf{C}oupled \textbf{A}ttribute–\textbf{T}opology Invariance Learning (\textbf{CATeye}) for generalizable voucher abuse detection. It tackles coupled attribute-topology shifts by explicitly isolating invariant context in both node attributes and graph structure. To achieve this, CATeye introduces an \emph{Attribute Invariance Selector} to predict node-adaptive binary masks to retain invariant attributes, and an \emph{Edge Invariance Selector} to extract an invariant subgraph conditioned on the selected attributes. It then constructs multiple views with different degrees of invariance and performs multi-view learning to emphasize stable predictive signals of voucher abuse while discouraging reliance on non-invariant views. Together, CATeye \textit{sees through} coupled shifts and achieves zero-shot generalization to unseen domains without fine-tuning.

Our contributions are summarized as follows:
\begin{itemize} [leftmargin=*]
    \item We identify coupled attribute-topology shifts in voucher abuse detection over time and across regions, and formulate the task under such domain shifts as a graph domain generalization problem. This formulation is realistic and well-aligned with practical deployment requirements, enabling zero-shot inference without retraining as new markets and time periods emerge.
    \item We propose a novel Coupled Attribute–Topology invariance learning framework (CATeye) that mitigates the unique coupled shifts in voucher abuse graphs. It jointly separates invariant and non-invariant components in both node attributes and graph structure, and learns invariant representations via multi-view objectives.
    \item We conduct extensive experiments on a real-world voucher abuse detection dataset from Lazada (covering a long time span and two countries in the Southeast Asian e-commerce market), as well as on the public Elliptic dataset. Empirical results demonstrate that CATeye consistently outperforms nine state-of-the-art baselines and yields up to an 8.61\% improvement in average F1 score over the strongest baseline.
\end{itemize}

\section{Related Work}

\noindent\textbf{Graph anomaly detection for voucher abuse detection. }
Voucher abuse detection has been studied using graph-based models that capture inter-order relations~\cite{wen2023voucher}. More broadly, graph anomaly detection (GAD) is widely applied in fraud detection~\cite{kim2024temporal}, intrusion detection~\cite{caville2022anomal}, and spam detection~\cite{yu2024gfd}. Most GAD methods implicitly assume training and test data follow the same distribution, which is often violated in practice when anomalous patterns evolve over time or vary across regions, causing severe distribution shifts~\cite{pan2025survey}. To improve robustness, recent works explore augmentation~\cite{zhou2023improving,ren2024heterophilic}, leverage target-domain data during training~\cite{chen2024towards,wang2023cross}, or perform test-time adaptation~\cite{wang2024goodat,zhengtest}. In this work, we make the minimal assumption that target-domain data are unavailable during training and focus on the more challenging domain generalization problem. Several studies~\cite{chen2024learning,fan2023generalizing} have shown that GNNs can overfit to variant correlations in the training data and thus degrade under distribution shift, requiring frequent retraining. To address this limitation, we aim to learn models that generalize effectively to future periods and unseen markets in a zero-shot manner.

\noindent\textbf{Graph domain generalization (DG).}
DG aims to train models on multiple source domains that generalize well to unseen target domains without fine-tuning~\cite{zhou2022domain}. Although many DG methods have been developed to handle distribution shifts in Euclidean data~\cite{arjovsky2019invariant,krueger2021out,ahuja2021invariance}, applying them directly to graph-structured data remains challenging due to the non-Euclidean nature of graphs~\cite{zhang2024survey}. 
Graph-level DG methods mainly extract invariant subgraphs~\cite{wu2022discovering,piao2024improving,sun2024dive} and learn invariant representations via disentanglement objectives~\cite{li2021disentangled,li2024disentangled}.
However, they tend to degrade significantly on node-level tasks. Recent attempts extend graph DG to node-level tasks by leveraging risk-extrapolation on virtual environments~\cite{wu2022handling}, pseudo-environment estimation~\cite{wu2024graph}, data-centric operations~\cite{zhao2025traci}, and meta-learning~\cite{tian2025mldgg}. 
Going beyond these efforts, we focus on voucher abuse graphs with coupled attribute-topology shifts, where attribute shifts induce shifted edges. Existing methods either treat attribute and structure as independent sources of variation, or assume invariance in the embedding space. Thus, they struggle with coupled shifts. To address this, we explicitly disentangle invariant and variant components in both node attributes and graph structure.

\begin{figure}[!t]
    \centering
    \includegraphics[width=\linewidth, trim={0 0 2.5cm 0},clip]{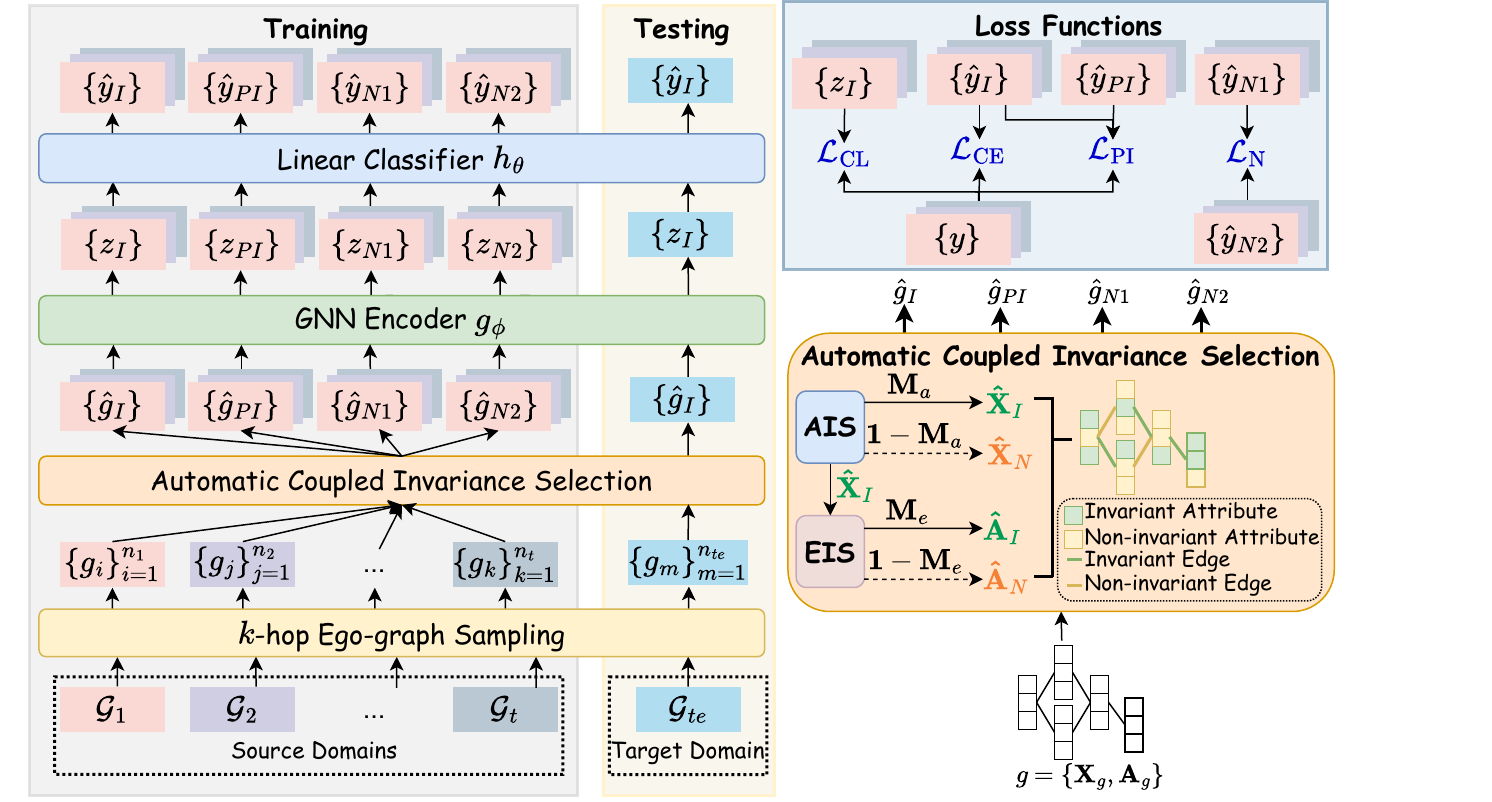}
    
    \caption{An overview of CATeye. \textbf{Left}: Training and testing pipeline. During training, we sample $k$-hop ego-graphs $g$ from multiple source graphs $\{\mathcal{G}_1,\dots,\mathcal{G}_t\}$. During inference, an unseen target graph $\mathcal{G}_{te}$ is evaluated without fine-tuning using only the invariant view $\hat{g}_I$. \textbf{Lower right}: Attribute Invariance Selector (AIS) and Edge Invariance Selector (EIS) are employed to automatically disentangle node attributes $\mathbf{X}_g$ and structure $\mathbf{A}_g$ into invariant and non-invariant parts. The resulting components form four views: invariant $\hat{g}_{I} = (\hat{\mathbf{X}}_I, \hat{\mathbf{A}}_I)$, partially invariant $\hat{g}_{PI} = (\hat{\mathbf{X}}_I, \hat{\mathbf{A}}_N)$, and non-invariant $\hat{g}_{N1} = (\hat{\mathbf{X}}_N, \hat{\mathbf{A}}_N)$ and $\hat{g}_{N2} = (\hat{\mathbf{X}}_N, \hat{\mathbf{A}}_I)$. \textbf{Upper right}:  The framework is optimized with multi-view objectives: $\mathcal{L}_\text{CL}$ and $\mathcal{L}_\text{CE}$ on $\hat{g}_I$, $\mathcal{L}_\text{PI}$ to suppress non-invariant structure in $\hat{g}_{PI}$, and entropy maximization $\mathcal{L}_\text{N}$ to discourage reliance on non-invariant views. Subscripts in the notations after the Automatic Coupled Invariance Selection layer are omitted for simplicity.}
    \label{fig:AGCID-overview}
\end{figure}

\section{Methodology}

\subsection{Problem Statement}
For each domain $s$, we construct an order graph $\mathcal{G}_s=(\mathcal{V}_s,\mathcal{E}_s)$ with node attributes $\mathbf{X}_s=[\mathbf{x}_1,\dots,\mathbf{x}_{|\mathcal{V}_s|}]^\top\in\mathbb{R}^{|\mathcal{V}_s|\times d}$, where $\mathcal{V}_s=\{v_1, v_2, \dots, v_{|\mathcal{V}_s|}\}$ represents the node set associated with $\mathbf{X}_s$, $\mathcal{E}_s$ is the edge set containing all the edges between orders, $|\mathcal{V}_s|$ is the number of orders in domain $s$, and $d$ is the number of attributes collected from buyer profiles and behavioral data. Labels are partially observed as $\mathcal{Y}_s=[y_1,\dots,y_{|\mathcal{V}_s|}]^\top\in\{0,1,-1\}^{|\mathcal{V}_s|}$, where $y_i=1$ denotes an abusive order, $y_i=0$ denotes a normal order, and $y_i=-1$ indicates an unlabeled order.

We are given a set of historical source graphs $\mathcal{G}_{tr}$, where $\mathcal{G}_{tr}=\{\mathcal{G}_1, \mathcal{G}_2, \dots, \mathcal{G}_t\}$ and each graph represents a distinct domain. Given an unseen target graph $\mathcal{G}_{te}=(\mathcal{V}_{te}, \mathcal{E}_{te})$ with node attributes $\mathbf{X}_{te}$, our goal is to \textbf{learn a model on $\mathcal{G}_{tr}$ that accurately predicts abuse on $\mathcal{G}_{te}$ without any further fine-tuning or adaptation}. 
Note that the node sets and structures of $\mathcal{G}_{tr}$ and $\mathcal{G}_{te}$ are disjoint, i.e., $\mathcal{G}_{tr} \cap \mathcal{G}_{te} = \emptyset$, and there exists a distribution shift between source and target graphs.

\noindent\textbf{Ego-Graph as instance.}
We sample $k$-hop ego-graphs from each source graph independently to build localized training instances.
For the $i$-th labeled node in a source graph $\mathcal{G}_s$, let $g_i = (\mathcal{V}_{g_i}, \mathcal{E}_{g_i})$ denote its $k$-hop ego-graph, where
\begin{equation}
\begin{split}
    \mathcal{V}_{g_i} = \{v|d(v_{g_i}, v) \le k \text{, and } v \in \mathcal{V}_s\},\\
    \mathcal{E}_{g_i} = \{(i,j)|(i,j) \in \mathcal{E}_s \text{, and } v_i, v_j \in \mathcal{V}_{g_i}\},
\end{split}
\end{equation}
$d(\cdot,\cdot)$ is the shortest path distance between two nodes, $v_{g_i}$ is the central node of $g_i$ and $k$ controls the neighborhood radius. This sampling procedure defines \textit{localized} instances for \textit{scalable} training and inference. Each instance is represented as $\{g_i=(\mathbf{X}_{g_i}, \mathbf{A}_{g_i}), y_{g_i}\}$, where $\mathbf{X}_{g_i} \in \mathbb{R}^{|\mathcal{V}_{g_i}| \times d}$ is the node attributes of $\mathcal{V}_{g_i}$, $\mathbf{A}_{g_i}$ is the adjacency matrix corresponding to $\mathcal{E}_{g_i}$, and $y_{g_i}$ is the label of the central node. Extracted ego-graphs are denoted as $G_s = [g_1, g_2, \dots, g_{n_s}]$, where $n_s$ is the number of labeled nodes in $\mathcal{G}_s$.

\subsection{Coupled Shifts in Graph Generation}

Prior graph DG works typically assume \textit{non-coupled} shifts, where attribute and structure shifts are independent, or assume invariance can be fully enforced in the embedding space, making them less effective at mitigating shifts inherent in voucher abuse graphs.
To understand the distribution shift in voucher abuse graphs, we introduce an illustrative model of the graph generation process in Figure~\ref{fig: scm}. This model clarifies how the effects of environment propagate and cause coupled attribute-topology shifts. 

The illustration includes an invariant factor $C$, an environment variable $E$, variant factors $S_X$ and $S_A$, node attributes $X$, graph structure $A$, and labels $Y$. Labels follow a stable mechanism $C \!\to\! Y$, while environment-dependent variations enter through two variant factors with $E\!\to\! S_X$ and $E\!\to\! S_A$. Observed node attributes and graph structure are generated by $\{C, S_X\} \to X$ and $\{X, S_A\} \to A$. 
The illustrative model shows how environment-driven variations in both attributes and structure lead to \textit{coupled attribute-topology shifts}. In particular, attribute shifts induce structural shifts because edges are constructed based on order attributes, causing coupled shifts across both attributes and structure. The mechanisms that characterize coupled shifts are highlighted by red arrows.

Motivated by this, we aim to train models that rely on invariant components in both attributes and structure while remaining insensitive to environment-specific variations, enabling robust generalization across time periods and regions. 

\subsection{Overview of CATeye}

Our approach addresses the multi-source graph domain generalization problem in voucher abuse detection using a shift-aware invariance perspective. We seek to mitigate coupled attribute-topology shifts by learning invariant order representations whose predictive relationship to abuse remains stable across different environments. The proposed Coupled Attribute–Topology invariance learning framework (CATeye) consists of two stages: (i) \textbf{Automatic Coupled Invariance Selection}, where the Attribute Invariance Selector (AIS) and Edge Invariance Selector (EIS) automatically disentangle node attributes and graph structure into invariant and non-invariant components, and (ii) \textbf{Multi-View Invariant Representation Learning}, where the decomposed components are combined into multiple views with different levels of invariance and optimized via view-specific objectives to emphasize invariance and suppress environment-specific variation. We present Automatic Coupled Invariance Selection in Section~\ref{sec:Decomposition} and Multi-View Invariant Representation Learning in Section~\ref{sec:Multi-View}. These two stages jointly enable zero-shot generalization to unseen time periods and markets. Figure~\ref{fig:AGCID-overview} provides an overview of CATeye. 

\subsection{Automatic Coupled Invariance Selection}
\label{sec:Decomposition}

To address coupled shifts, CATeye explicitly disentangles invariant and non-invariant information in both attributes and structure. For each sampled ego-graph, two specialized invariance selection modules, the \textit{Attribute Invariance Selector (AIS)} and the \textit{Edge Invariance Selector (EIS)}, automatically and adaptively extract its invariant attributes and structure. We describe each module in detail below.

\noindent\textbf{Attribute Invariance Selector (AIS). }
CATeye mitigates domain-dependent redundancy in node attributes via a learnable Attribute Invariance Selector. Given a node feature $\mathbf{x}\in\mathbb{R}^d$, the selector $h_{\theta_x}$ outputs \textit{node-adaptive} scores $\mathbf{p}_a = [p_a^1, p_a^2, \dots, p_a^d]^{\top} = h_{\theta_x}(\mathbf{x}) \in \mathbb{R}^d$. This vector is then transformed into a binary mask $\mathbf{m}_a\in\{0,1\}^d$ using Gumbel-Softmax reparameterization~\cite{maddison2016concrete}, i.e., $m_a^i=\text{Gumbel-Softmax}(p_a^i)$, where $i \in \{1, 2, ..., d\}$. 
Each element determines whether the corresponding attribute is retained ($m_a^i=1$) or discarded ($m_a^i=0$). The Gumbel-Softmax is defined as follows:
\begin{equation}
    \resizebox{.96\linewidth}{!}{$
    \displaystyle
     \text{Gumbel-Softmax}(p)= 
    \sigma\left(\frac{\log \epsilon-\log (1-\epsilon)+\log \frac{\sigma(p)}{1-\sigma(p)}}{\tau_w}\right)
    $},
\end{equation}
where $\sigma(\cdot)$ is the sigmoid function, $\epsilon \sim \operatorname{Uniform}(0,1)$, and $\tau_w$ is the temperature controlling the discretization. As $\tau_w \rightarrow 0$, it approximates the Bernoulli distribution parameterized by $p$. The reparameterization ensures differentiable binary discretization. 
For an ego-graph $g$, stacking node masks yields $\mathbf{M}_a = [\mathbf{m}_a^1, \mathbf{m}_a^2, ..., \mathbf{m}_a^{|\mathcal{V}_g|}]^{T} \in \{0, 1\}^{|\mathcal{V}_g| \times d}$. The invariant attributes $\hat{\mathbf{X}}_I$ and non-invariant attributes $\hat{\mathbf{X}}_N$ are derived as:
\begin{equation}
    \hat{\mathbf{X}}_I = \mathbf{M}_a \odot \mathbf{X}_g, \qquad
    \hat{\mathbf{X}}_N = (\mathbf{1}_{|\mathcal{V}_g| \times d} - \mathbf{M}_a) \odot \mathbf{X}_g.
\end{equation}
Here, $\hat{\mathbf{X}}_I$ contains invariant attributes that are
stable across domains, while $\hat{\mathbf{X}}_N$ captures non-invariant features that fail to generalize.

\noindent\textbf{Edge Invariance Selector (EIS). }
Because the graph structure is induced by attribute proximity, non-invariant attributes and coincidental overlaps inevitably lead to non-invariant edges. Including these connections in message passing amplifies domain-specific information and degrades generalization. To mitigate this negative effect, CATeye employs the Edge Invariance Selector (EIS) to sample an invariant subgraph. EIS consists of a GNN $g_{\phi_e}$ and an MLP $h_{\theta_e}$. It first computes node embeddings $\mathbf{H} =[\mathbf{h}_1, \mathbf{h}_2, ..., \mathbf{h}_{|\mathcal{V}_g|}]^{T}=g_{\phi_e}(\hat{\mathbf{X}}_I, \mathbf{A}_g)$, then scores each edge $e_{i,j}$ by $p_e^{ij} = h_{\theta_e}([\mathbf{h}_i; \mathbf{h}_j])$, where $\mathbf{h}_i, \mathbf{h}_j \in \mathbf{H}$, and $[\cdot;\cdot]$ is the concatenation operation. A binary edge mask is then derived as $\mathbf{M}_e^{ij}=\text{Gumbel-Softmax}(p_e^{ij})$.
An edge from node $v_i$ to node $v_j$ remains in the invariant subgraph if $\mathbf{M}_e^{ij}=1$ and is dropped otherwise. 
The resulting invariant structure $\hat{\mathbf{A}}_I$ and non-invariant structure $\hat{\mathbf{A}}_N$ are:
\begin{equation}
    \hat{\mathbf{A}}_I = \mathbf{M}_e \odot \mathbf{A}_g, \quad 
\hat{\mathbf{A}}_N = (\mathbf{1}_{|\mathcal{V}_g| \times |\mathcal{V}_g|} - \mathbf{M}_e) \odot \mathbf{A}_g.
\end{equation}
Computing edge importance from $\hat{\mathbf{X}}_I$ rather than from the original $\mathbf{X}_g$ keeps the invariant structure from being disturbed by non-invariant attributes.

\subsection{Multi-View Invariant Representation Learning}
\label{sec:Multi-View}

\noindent\textbf{View construction. }
For each ego-graph $g$, we generate four views by combining invariant and non-invariant components: (1) \textit{Invariant View} $\hat{g}_I = (\hat{\mathbf{X}}_I, \hat{\mathbf{A}}_I)$, which provides fully invariant context; (2) \textit{Partially Invariant View} $\hat{g}_{PI} = (\hat{\mathbf{X}}_I, \hat{\mathbf{A}}_N)$, in which the invariant attributes carry information transferable to unseen domains, although the structure is unreliable for generalization; and (3) \textit{Non-Invariant Views}: $\hat{g}_{N1} = (\hat{\mathbf{X}}_N, \hat{\mathbf{A}}_N)$ and $\hat{g}_{N2} = (\hat{\mathbf{X}}_N, \hat{\mathbf{A}}_I)$, which include non-invariant attributes. With invariant structure but non-invariant attributes, $\hat{g}_{N2}$ is still considered non-invariant because structure contributes to invariance only when it connects invariant attributes. 

\subsubsection{Optimization Objectives}

For each source graph $\mathcal{G}_s \in \mathcal{G}_{tr}$, we sample labeled ego-graphs $G_s = [g_1, g_2, \dots, g_{n_s}]$ with labels $\mathbf{y}_s=[y_{g_1}, y_{g_2}, ..., y_{g_{n_s}}]$. Each ego-graph $g_i$ is decomposed into four views $\hat{g}_{i,v}$, where $v \in \{I, PI, N1, N2\}$. A shared GNN encoder $g_{\phi}$ produces representations $z_{i,v} = g_{\phi}(\hat{g}_{i,v}) \in \mathbb{R}^{d_h}$, where $d_h$ denotes the hidden size, followed by a linear classifier $h_{\theta}$ producing prediction $\hat{y}_{i,v} = h_{\theta}(z_{i,v}) \in \mathbb{R}^{2}$. We optimize the framework with view-specific objectives designed to encourage invariance and suppress variant shortcuts.

\noindent\textbf{Invariant View. }
It should produce a decisive prediction for the label, and remain domain-invariant, i.e., $\max_{\Theta} \ \text{I}(\hat{g}_{i, I}; y_{g_i}) \text{, s.t. } \hat{g}_{i, I} \indep E$, where $\Theta = \{\theta_x, \phi_e, \theta_e, \phi, \theta\}$ denotes all learnable parameters, $\text{I}(\cdot;\cdot)$ measures the mutual information between two random variables, and $E$ denotes the domain label. We implement this with:
\begin{equation}
    \mathcal{L}_\text{I} = 
    \underbrace{\mathcal{L}_{\text{CE}}(\hat{y}_{i, I}, y_{g_i})}_{\text{maximize } \text{I}(\hat{g}_{i, I}; y_{g_i})}
    + \omega_{\text{CL}}
    \underbrace{\mathcal{L}_{\text{CL}}(\hat{g}_{i, I})}_{\text{enforce } \hat{g}_{i, I} \indep E},
\end{equation}
where $\mathcal{L}_{\text{CE}}$ is the cross-entropy loss, and $\mathcal{L}_{\text{CL}}$ is an invariance-inducing contrastive loss. $\mathcal{L}_{\text{CL}}$ is defined as:
\begin{equation}
    \mathcal{L}_{\text{CL}}(\hat{g}_{i, I})=\underset{m \in \mathcal{P}(i)}{\mathbb{E}}\Big[-\log \frac{\exp ({z_{i, I}}^{\top} {z_{m, I}} / \tau)}{\underset{k \in \mathcal{P}(i) \cup \mathcal{N}(i)}{\mathbb{E}} [\exp ({z_{i, I}}^{\top} z_{k, I} / \tau)]}\Big],
\end{equation}
where $\mathcal{P}(i) = \{m: y_{g_m}=y_{g_i}, d_{g_m}\neq d_{g_i}\}$ is the set of positives (same class, different domain) and $\mathcal{N}(i) = \{n: y_{g_n}\neq y_{g_i}\}$ is the set of negatives. $\tau$ is a temperature hyperparameter. $\mathcal{L}_{\text{CL}}$ encourages invariant embeddings by pulling samples from the same class but different domains to be close. Aggregating same-class pairs across \textit{all} source domains greatly increases the effective supervision signal, making $\mathcal{L}_{\text{CL}}$ effective even under severe label scarcity in voucher abuse detection.

\noindent\textbf{Partially Invariant View.}
It should be decisive about the label but less informative compared to the Invariant View, because of its non-invariant structure. We therefore optimize it only when it underperforms the Invariant View, i.e., $\max_{\Theta} \ \text{I}(\hat{g}_{i, PI}; y_{g_i}) \text{, s.t. } \text{I}(\hat{g}_{i, I}; y_{g_i}) > \text{I}(\hat{g}_{i, PI}; y_{g_i})$. 
The practical implementation is as follows:
\begin{equation}
    \mathcal{L}_\text{PI} = \mathcal{L}_{\text{CE}}(\hat{y}_{i, PI}, y_{g_i}) \cdot \mathbbm{1} \Big(\mathcal{L}_{\text{CE}}(\hat{y}_{i, PI}, y_{g_i}) \ge \mathcal{L}_{\text{CE}}(\hat{y}_{i, I}, y_{g_i})\Big).
\end{equation}
It prevents the model from relying on non-invariant structure.

\noindent\textbf{Non-invariant Views.} 
To discourage reliance on $\hat{g}_{i,N1}$ and $\hat{g}_{i,N2}$, i.e., $\min_{\Theta} \ \text{I}(\hat{g}_{i, N1}; y_{g_i}) + \text{I}(\hat{g}_{i, N2}; y_{g_i})$, we maximize prediction entropy as follows:
\begin{equation}
    \mathcal{L}_\text{N} = -\frac{1}{2}[
    \text{H}(\hat{y}_{i, N1}) + \text{H}(\hat{y}_{i, N2})
    ],
\end{equation}
where $\text{H}(\cdot)$ is Shannon entropy. This forces the classifier to remain uncertain on Non-invariant Views. Compared to minimizing the negative cross-entropy, entropy maximization avoids overfitting to the limited available labels, which is particularly beneficial in label-scarce scenarios of the voucher abuse detection task.

\noindent\textbf{Overall training objective.}  
The final optimization objective combines all view-specific losses:
\begin{equation}
\min_{\Theta} \ \mathcal{L}_\text{I} + \omega_{\text{PI}} \cdot  \mathcal{L}_\text{PI} + \omega_{\text{N}} \cdot \mathcal{L}_\text{N},
\end{equation}
where $\omega_{\text{PI}}$ and $\omega_{\text{N}}$ balance the partially invariant and non-invariant view objectives. Collectively, the overall objective guides the AIS and EIS to automatically disentangle invariant and non-invariant information from both attributes and structure. It encourages the encoder to learn invariant representations and suppress environment-dependent correlations. 

\section{Experiments}

\subsection{Datasets}
We evaluate our method on datasets that exhibit natural domain shifts and severe class imbalance, which are challenging for graph domain generalization.

\noindent\textbf{Voucher Abuse Detection Dataset. }
We build a proprietary dataset from an e-commerce platform, Lazada, to study voucher abuse detection under distribution shift. The \textit{source domains} consist of ID0501--ID0505, i.e., order graphs from the Indonesia market collected on 1--5 May 2025. More details regarding graph construction are presented in Section~\ref{sec:graph-construction}.
The \textit{target domains} include 16 order graphs from both \textit{Indonesia and Vietnam}, covering both \textit{promotional campaign days and regular days}. The most recent target graph is collected on 25 August 2025, which introduces a nearly four-month temporal gap relative to the source data. This dataset captures domain shifts from both \textbf{time} (evolving abusive patterns) and \textbf{region} (cross-country behavioral differences). Abusive orders account for only a tiny fraction of all orders, making this a highly imbalanced task.

\noindent\textbf{Elliptic Dataset.}
The Elliptic dataset\footnote{https://www.kaggle.com/datasets/ellipticco/elliptic-data-set}~\cite{weber2019anti} is a public benchmark of temporal Bitcoin transaction graphs collected evenly with an interval of about two weeks. Nodes represent transactions, and edges represent Bitcoin flows. The task is to detect illicit transactions under temporal distribution shift and class imbalance.

\subsection{Graph Construction for the Voucher Abuse Detection Dataset}
\label{sec:graph-construction}

We construct one order graph for each day in each region. In each graph, nodes represent individual orders, with $d=801$ attributes from (i) user profiles (e.g., region, account age), (ii) transaction records (e.g., payment method, purchase amount, timestamp), and (iii) click-path embeddings that encode sequences of user actions using an unsupervised skip-gram model FastText~\cite{bojanowski2017enriching}. 
Graph structure is constructed to capture connections between orders that indicate potential abuse. Two nodes are connected if they satisfy both of the following conditions: (1) \textit{profile proximity} (shared identifiers such as shipping address) and (2) \textit{behavior proximity} (overlapping transactional patterns such as near-synchronous checkout and similar sellers or product categories), consistent with prior findings~\cite{cao2014uncovering} that abuse often involves loosely synchronized actions across shared entities. 
Order labels (abusive or legitimate) are generated using rules curated by domain experts. Because only a small fraction of orders can be annotated with high confidence, labeled training data remains \textit{limited}.

\subsection{Baselines}
Besides Empirical Risk Minimization (ERM)~\cite{vapnik1991principles}, we compare CATeye against three classes of state-of-the-art baselines. 
\begin{itemize} [leftmargin=*]
    \item Graph anomaly detection method with enhanced generalization: AugAN~\cite{zhou2023improving} enriches the training data through data augmentation and uses episodic training to learn from the augmented samples.
    \item General-purpose domain generalization baselines: IRM~\cite{arjovsky2019invariant}, IB-IRM~\cite{ahuja2021invariance}, V-REx~\cite{krueger2021out}, and DANN~\cite{ganin2016domain}. These methods aim to learn domain-invariant predictors across multiple domains.
    \item Graph domain generalization baselines: TRACI~\cite{zhao2025traci}, EERM~\cite{wu2022handling}, and CaNet~\cite{wu2024graph}. TRACI improves generalization via data-centric operations, whereas EERM and CaNet enforce invariance in the embedding space. 
\end{itemize}

\begin{table*}[!t]
\centering
\caption{The F1 scores (\%) on the voucher abuse detection dataset. The best performance is highlighted in \textbf{bold}, and the second-best performance is {\ul underlined}. Results from \colorbox{blue!20}{campaign days} and \colorbox{green!20}{regular days} are distinguished by different background colors.}
\resizebox{0.9\textwidth}{!}{
\begin{tabular}{c|c|ccccc|ccc|c}
\toprule
\begin{tabular}[c]{@{}c@{}}Target\\ Domain\end{tabular}              & AugAN      & ERM              & IRM        & IB-IRM     & V-REx      & DANN       & TRACI        & EERM             & CaNet             & \begin{tabular}[c]{@{}c@{}}\textbf{CATeye}\\ \textbf{(Ours)}\end{tabular}        \\ \midrule
\cellcolor{blue!20}ID0605  & 66.14±6.27 & 86.11±0.40       & 86.58±0.92 & 86.66±0.64 & 86.67±0.90 & 77.14±2.17 & 81.16±0.34  & {\ul 87.06±1.22} & 85.67±1.04        & \textbf{88.05±1.05} \\
\cellcolor{blue!20}ID0606  & 66.61±5.82 & 81.68±1.42       & 82.47±1.55 & 83.48±1.92 & 82.86±0.80 & 68.63±2.61 & 73.10±0.28  & {\ul 83.48±1.30} & 81.28±1.61        & \textbf{84.42±2.14} \\
\cellcolor{blue!20}ID0707  & 60.64±5.89 & 78.15±0.85       & 77.72±1.30 & 76.83±1.12 & 78.07±1.16 & 66.13±5.79 & 76.41±1.73 & {\ul 80.76±2.25} & 77.52±1.62        & \textbf{83.24±1.41} \\
\cellcolor{blue!20}ID0808  & 55.08±3.82 & 76.42±2.12       & 75.68±2.17 & 76.31±1.75 & 75.91±2.45 & 61.12±4.67 & 70.11±1.88  & {\ul 79.09±2.28} & 75.23±3.22        & \textbf{82.38±1.94} \\
\cellcolor{green!20}ID0625 & 60.71±4.03 & 76.97±0.94       & 75.98±1.02 & 74.07±0.50 & 76.75±1.28 & 63.46±8.22 & 76.70±1.84 & {\ul 79.47±2.11} & 77.41±2.54        & \textbf{81.20±1.37}  \\
\cellcolor{green!20}ID0725 & 60.88±4.29 & 76.96±1.77       & 75.57±2.00 & 74.50±1.54 & 77.09±1.41 & 61.25±9.32 & 77.19±2.52 & 78.99±2.00       & {\ul 79.46±2.94}  & \textbf{81.33±2.18} \\
\cellcolor{green!20}ID0825 & 61.86±4.44 & 79.96±1.44       & 79.10±1.95 & 78.74±1.51 & 79.69±1.81 & 64.44±6.19 & 75.90±1.49 & {\ul 82.17±1.62} & 79.88±1.54        & \textbf{84.57±1.34} \\ \midrule
\cellcolor{blue!20}VN0605  & 58.14±5.67 & {\ul 74.56±2.92} & 73.15±3.51 & 71.73±3.67 & 73.91±3.39 & 67.32±6.60 & 73.43±1.42 & 74.40±3.81       & 70.59±6.22        & \textbf{77.56±2.85} \\
\cellcolor{blue!20}VN0606  & 56.92±4.13 & {\ul 73.58±3.74} & 71.09±4.05 & 69.10±4.58 & 73.07±3.58 & 64.22±3.28 & 72.10±1.56  & 70.65±4.84       & 70.21±6.63        & \textbf{76.01±2.71} \\
\cellcolor{blue!20}VN0607  & 66.65±4.99 & 76.83±1.92       & 75.77±2.29 & 74.79±2.85 & 76.68±2.45 & 68.47±3.90 & 76.32±1.74  & 76.51±2.92       & {\ul 76.96±4.38}  & \textbf{79.36±1.95} \\
\cellcolor{blue!20}VN0608  & 68.30±5.89 & 71.51±3.43       & 69.67±4.36 & 67.33±4.91 & 70.96±3.20 & 64.47±4.22 & {\ul 72.83±2.46}  & 68.83±4.14       & 72.01±8.49  & \textbf{73.02±3.34} \\
\cellcolor{blue!20}VN0707  & 70.67±2.96 & 83.71±1.06       & 82.88±1.50 & 81.75±1.54 & 83.26±1.27 & 72.09±2.53 & 80.60±1.52  & 83.41±2.12       & {\ul 84.10±1.38}  & \textbf{86.09±1.60} \\
\cellcolor{blue!20}VN0808  & 64.67±4.68 & {\ul 82.39±1.00} & 81.31±0.90 & 80.06±0.87 & 82.11±1.09 & 68.81±2.63 & 76.98±1.36 & 82.09±2.12       & 81.54±1.24        & \textbf{85.06±1.33} \\
\cellcolor{green!20}VN0625 & 67.93±4.59 & 71.05±4.34       & 69.66±4.63 & 67.24±4.93 & 70.90±4.73 & 65.13±4.61 & 72.39±2.26  & 68.88±4.53       & {\ul 72.59±5.95}  & \textbf{73.15±3.03} \\
\cellcolor{green!20}VN0725 & 70.09±4.62 & 81.93±1.90       & 80.78±2.06 & 79.22±0.97 & 81.72±1.77 & 72.56±3.73 & 80.34±2.58  & 79.82±2.69       & {\ul 82.68±4.13}  & \textbf{83.45±1.53} \\
\cellcolor{green!20}VN0825 & 67.30±8.73 & 81.43±1.53       & 80.63±1.96 & 79.46±2.16 & 81.28±1.69 & 72.94±3.08 & 81.05±1.61  & 81.36±2.30       & {\ul 82.15±3.96}  & \textbf{82.95±1.48} \\ \midrule
Avg.                       & 63.91      & 78.33            & 77.38      & 76.33      & 78.18      & 67.39      & 76.04       & {\ul 78.56}      & 78.08             & \textbf{81.37}     \\
\bottomrule
\end{tabular}
}
\label{Table:result-voucher}
\end{table*}

\begin{table*}[!t]
\centering
\caption{The F1 scores (\%) on the Elliptic dataset. The best performance is highlighted in \textbf{bold}, and the second-best performance is {\ul underlined}.}
\resizebox{0.9\textwidth}{!}{
\begin{tabular}{c|c|ccccc|ccc|c}
\toprule
\begin{tabular}[c]{@{}c@{}}Target\\ Domain\end{tabular} & AugAN      & ERM        & IRM              & IB-IRM      & V-REx      & DANN       & TRACI        & EERM             & CaNet               & \begin{tabular}[c]{@{}c@{}}\textbf{CATeye}\\ \textbf{(Ours)}\end{tabular}         \\\midrule
T1            & 65.47±1.52 & 92.73±0.48 & 92.04±0.26       & 93.55±0.40   & 92.61±0.36 & 83.95±1.08 & 92.07±0.46  & 89.51±0.63       & \textbf{95.13±0.83} & {\ul 94.20±1.31}    \\
T2            & 57.42±1.05 & 78.37±3.01 & 79.17±1.81       & 68.46±8.33  & 79.80±1.74  & 73.86±3.85 & 71.99±6.87  & {\ul 84.43±0.53} & 83.73±6.21          & \textbf{89.74±0.85} \\
T3            & 42.55±1.98 & 62.34±5.01 & 64.58±4.00        & 44.64±14.95 & 63.71±4.89 & 56.53±8.24 & 52.03±9.41  & {\ul 70.30±0.86}  & 67.22±11.83         & \textbf{76.47±4.95} \\
T4            & 28.04±7.04 & 54.41±4.56 & 57.76±3.74       & 40.33±15.19 & 56.04±4.43 & 45.77±6.63 & 37.20±11.90 & {\ul 60.00±1.19}  & 43.03±18.41         & \textbf{68.98±2.04} \\
T5            & 33.01±1.17 & 58.76±2.70  & {\ul 60.18±2.12} & 48.34±10.90  & 59.89±2.36 & 48.15±4.23 & 41.16±11.41 & 59.72±0.41       & 49.03±12.80          & \textbf{74.04±2.89} \\
T6            & 41.44±1.83 & 60.88±6.41 & {\ul 64.95±5.58} & 49.40±16.31  & 64.46±4.24 & 45.33±4.95 & 41.03±12.00 & 63.76±1.02       & 51.15±18.27         & \textbf{71.24±4.87} \\
T7            & 42.35±3.95 & 67.13±1.56 & {\ul 70.64±4.30}  & 64.72±6.71  & 67.73±6.77 & 30.75±3.40  & 56.09±13.77  & 54.48±2.44       & 56.64±15.65         & \textbf{79.17±4.70} \\
T8            & 29.36±4.06 & 52.77±2.28 & {\ul 56.49±4.47} & 44.22±8.08  & 55.15±4.19 & 26.94±3.58 & 28.77±13.85 & 50.04±1.87       & 36.48±20.85         & \textbf{60.84±11.27} \\\midrule
Avg.          & 42.46      & 65.92      & {\ul 68.23}      & 56.71       & 67.42      & 51.41      & 52.54       & 66.53            & 60.30               & \textbf{76.84}  \\
\bottomrule
\end{tabular}
}
\label{Table:result-elliptic}
\end{table*}

\subsection{Implementation Details}
All models are implemented in PyTorch~\cite{paszke2019pytorch} and trained on a single NVIDIA A100 GPU with 80 GB memory. For CATeye, we use a 2-layer GraphSAGE encoder~\cite{hamilton2017inductive} as $g_{\phi}$ with hidden dimension $d_h=128$ and ReLU activation function, followed by a linear classifier $h_{\theta}$. The AIS $h_{\theta_x}$ is implemented as a two-layer MLP with a hidden size $128$, and the EIS consists of a two-layer GNN $g_{\phi_e}$ followed by a two-layer MLP $h_{\theta_e}$ with the same hidden size. 
To avoid mask collapse, after generating $\mathbf{M}_a$ and $\mathbf{M}_e$, we retained only the top $\rho_a$ fraction of attributes and the top $\rho_e$ fraction of edges by score, with $\rho_a=\rho_e=0.5$. 
An ablation study is conducted in Section~\ref{sec:ablation-sens_rho} to analyze the impact of these two parameters on model performance.
For baselines, we follow their official implementations with recommended hyperparameters. For DANN, which is originally designed for domain adaptation, we apply its domain-adversarial objective only on the source domains to align with the DG setting. 

For the voucher abuse detection dataset, we split the labeled data from ID0501--ID0505 into training and validation sets with equal sizes. The evaluation is conducted on 16 target-domain graphs. To reflect the label scarcity of real-world voucher abuse detection, we use only 50 anomalous nodes and 300 normal nodes per source domain for training, following the actual anomaly rate.  
For the Elliptic dataset, we use snapshots 6–10 for training and validation, split with a ratio of $60\%:40\%$, and snapshots 11–43 for testing. To account for the temporal shift, we group the 32 test snapshots into 8 chronological folds, named T1--T8, each containing 4 consecutive snapshots. 

We perform grid search on the hyperparameters as follows: $\omega_\text{CL}, \omega_\text{PI}, \omega_\text{N} \in \{0.03, 0.1, 0.3, 1.0, 3.0\}$. The temperature parameters, $\tau_w$ for discretization and $\tau$ for the domain-invariant contrastive loss, are set to $0.2$ and $0.1$, respectively. 

During inference, we make predictions using only the invariant views of the test ego-graphs, making the model compatible with streaming production pipelines without requiring multi-graph context or retraining as new daily graphs arrive. CATeye supports zero-shot deployment to new markets and campaign periods as they emerge. We report the mean and standard deviation of F1 scores over 5 random seeds.

\subsection{Main Results}

The overall results on the voucher abuse detection dataset are presented in Table~\ref{Table:result-voucher}. CATeye consistently achieves the highest performance across all target domains, with an overall average F1 score of 81.37\%, outperforming the strongest baseline by 2.81\%.
Since all methods share identical constructed graphs, these gains stem entirely from model design.
Compared with general-purpose DG methods, i.e., IRM, IB-IRM, V-REx, and DANN, CATeye yields larger and more consistent improvements. The performance gain comes from jointly extracting invariant information from both attributes and structure, rather than learning invariance only in the hidden space. Notably, the improvements are most evident on campaign days, where coincidental overlaps in purchases introduce more unstable edges and variant correlations. By effectively filtering out such information, CATeye reduces the influence of environment-dependent signals and maintains robust performance in detecting abusive behavior.

Among graph DG baselines, TRACI underperforms several strong baselines, suggesting its data-centric transformations fail to address coupled shifts. EERM and CaNet, which enforce embedding-level invariance,  occasionally achieve second-best on individual domains but lack consistency. They still fall 2.81\% and 3.29\% short of CATeye, respectively, providing direct empirical evidence that invariance in the embedding space alone cannot resolve coupled attribute–topology shifts. 
ERM performs competitively when shifts are mild or variant correlations remain predictive, but degrades unpredictably under stronger shifts due to the reliance on unstable signals. Graph anomaly detection methods such as AugAN provide limited improvements, indicating that data augmentation alone is insufficient for generalization under domain shifts.

Evaluation on the Elliptic dataset, recorded in Table~\ref{Table:result-elliptic}, further highlights the enhanced generalization ability of our approach. CATeye achieves the best average F1 of 76.84\%, exceeding the second-best method by 8.61\% and delivering consistent improvements on later test folds, i.e., T4--T7, where temporal shift is larger. 

In summary, the results on both datasets demonstrate that CATeye not only generalizes effectively across time and regions in voucher abuse detection but also performs strongly on a public benchmark with temporal domain shifts. CATeye provides an effective and broadly applicable solution for graph anomaly detection under distribution shift.

\subsection{Ablation Study}

\begin{table}[!t]
\centering
\caption{Effects of the learning objectives for the views, and the two invariance selectors AIS and EIS. Results are average F1 scores (\%) over 16 target domains of the voucher abuse detection dataset. The number in the bracket illustrates the performance gap with CATeye. }
\resizebox{0.42\textwidth}{!}{
\begin{tabular}{c|c}
\toprule
Variants                                      & Avg. \\
\midrule
w/o $\mathcal{L}_\text{CL}$ on the invariant view   & 80.85 (-0.52)                     \\
w/o $\mathcal{L}_\text{PI}$ on the partially invariant view & 75.91 (-5.46)                    \\
w/o $\mathcal{L}_\text{N}$ on non-invariant views     & 80.30 (-1.07)                   \\
\midrule
stochastic attribute selection with random $\mathbf{M}_a$              & 74.87 (-6.50)                  \\
stochastic edge selection with random $\mathbf{M}_e$                   & 80.35 (-1.02)                 \\
\textcolor{black}{EIS conditioned on original $\mathbf{X}$ (not $\hat{\mathbf{X}}_I$)} & \textcolor{black}{80.91 (-0.46)}  \\
both selections are random                                & 73.62 (-7.75)                  \\
\midrule
\textbf{CATeye (Ours)}                                & \textbf{81.37}       \\   
\bottomrule
\end{tabular}
}
\label{Table:component-ablation}
\end{table}

\noindent\textbf{Effect of each component.}
We conduct ablation studies to evaluate the effect of the learning objectives for each view and the two invariance selectors. Results are presented in Table~\ref{Table:component-ablation}. Disabling any view-specific objective degrades performance, confirming that each loss term plays a distinct role in supporting generalization. In particular, $\mathcal{L}_\text{PI}$ yields the largest enhancement of 5.46\%, indicating that constraining the partially invariant view is crucial to prevent non-invariant structure from contaminating generalization. $\mathcal{L}_\text{CL}$ and $\mathcal{L}_\text{N}$ complement the objectives, showing benefits from strengthening invariant representations and discouraging reliance on non-invariant views. 
We further replace the learned masks in AIS and EIS with random binary masks. The results demonstrate that both selectors are critical. Randomizing AIS drastically reduces the performance by 6.50\%, highlighting the importance of identifying invariant attributes as a complement to invariant structure. Randomizing EIS causes a smaller but still notable degradation of 1.02\%, showing that refining the graph structure is useful, though the model is less sensitive to structural shortcuts than to attribute shortcuts. 
Replacing AIS-filtered $\hat{\mathbf{X}}_I$ with original $\mathbf{X}$ as EIS input drops F1 by 0.46\%, demonstrating that AIS captures genuinely invariant attributes and $\hat{\mathbf{X}}_I$ provides more reliable input for EIS. 
Randomizing both produces the largest performance decline by 7.75\%. These results highlight that identifying invariant attributes is especially important, and jointly selecting invariant attributes and edges yields the best generalization against coupled shifts.

\begin{figure}[!t]
    \centering
    \includegraphics[width=0.42\textwidth]{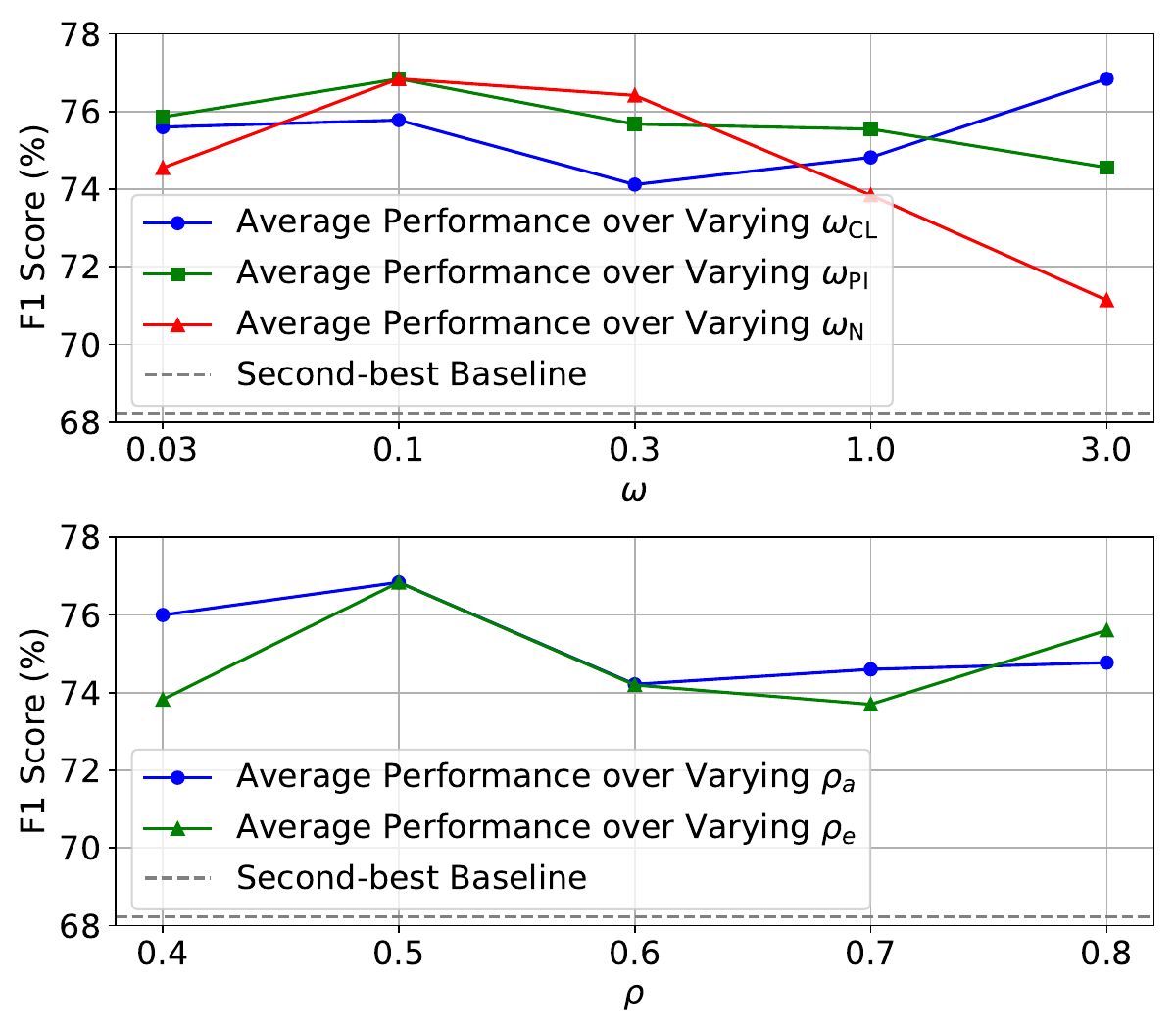}
    \caption{The plots show the average F1 scores (\%) of CATeye across eight test folds of the Elliptic dataset under varying hyperparameters. \textbf{Top}: Effects of $\omega_\text{CL}$, $\omega_\text{PI}$, and $\omega_\text{N}$, which control the loss weights for different views. \textbf{Bottom}: Effects of $\rho_a$ and $\rho_e$, which determine the proportion of unmasked invariant attributes and edges retained by the AIS and EIS. The grey dashed line indicates the performance of the second-best baseline for comparison.}
    \label{fig:sensitivity-all-ablation}
\end{figure}

\noindent\textbf{Sensitivities to hyperparameters.}
\label{sec:ablation-sens_rho}
The upper plot in Figure~\ref{fig:sensitivity-all-ablation} reports the average F1 scores of CATeye across eight test folds of the Elliptic dataset under varying values of hyperparameters $\omega_\text{CL}$, $\omega_\text{PI}$, and $\omega_\text{N}$, respectively. The grey dashed line shows the performance of the second-best baseline for reference. Overall, CATeye consistently outperforms the second-best baseline by a large margin and exhibits stable performance across different values of parameters. This indicates that our method remains robust to the configuration of hyperparameters.
The lower plot examines the sensitivity to $\rho_a$ and $\rho_e$, which control the proportion of unmasked invariant attributes and edges, respectively. The results show that CATeye remains stable across a broad range of unmask ratios, with performance consistently above the second-best baseline. This indicates that the model effectively leverages both attribute-level and structure-level manipulations without requiring exhaustive tuning of the masking ratios.

\section{Conclusion}
In this paper, we present the Coupled Attribute–Topology invariance learning (CATeye) for voucher abuse detection under domain shift. To \textit{see through} coupled shifts in voucher abuse graphs, CATeye automatically extracts invariant information from both node attributes and graph structure, and applies multi-view objectives to emphasize invariant representations and mitigate reliance on environment-specific variations. Extensive experiments on a real-world voucher abuse dataset and the public Elliptic benchmark show that CATeye consistently outperforms nine strong baselines and achieves robust zero-shot generalization to unseen domains.

\begin{acks}
    This research is partly supported by Alibaba Group and Alibaba-NTU Singapore Joint Research Institute (JRI), Nanyang Technological University, Singapore. It is also supported by the RIE2025 Industry Alignment Fund – Industry Collaboration Projects (IAF-ICP) (Award I2301E0026), administered by A*STAR, as well as supported by Alibaba Group and NTU Singapore through Alibaba-NTU Global e-Sustainability CorpLab (ANGEL).
\end{acks}

\section{GenAI Usage Disclosure}
We used large language models only to polish the writing and improve the clarity of the manuscript. All research ideas, methods, experiments, analyses, and conclusions were developed and verified by the authors.


\bibliographystyle{ACM-Reference-Format}
\bibliography{sample-base}

\end{document}